\pdfoutput=1
\documentclass[11pt]{article}

\usepackage[]{ACL2023}

\usepackage{times}
\usepackage{latexsym}

\usepackage[T1]{fontenc}
\usepackage[utf8]{inputenc}

\usepackage{microtype}

\usepackage{inconsolata}

\usepackage{amsmath}
\usepackage{amssymb}
\usepackage{amsfonts}
\usepackage{xcolor} 
\usepackage{enumitem}
\usepackage{graphicx}
\usepackage{booktabs}
\usepackage{tabularx}
\usepackage{multirow}
\usepackage{makecell}
\usepackage{utfsym}
\usepackage{stfloats}
\usepackage{algorithm}
\usepackage{algorithmic}
\graphicspath{{Figure/}}
\newcommand{\M}{QKConv}
\newcommand{\start}[1]{\vspace{.6mm}\noindent{{\bf #1}\ }}
\definecolor{my_blue}{RGB}{77, 115, 190}

\title{Query Enhanced Knowledge-Intensive Conversation via\\ Unsupervised Joint Modeling}

\author{Mingzhu Cai \quad Siqi Bao \quad Xin Tian \quad Huang He \quad Fan Wang \quad Hua Wu \\
        Baidu Inc., China \\
        \texttt{\{caimingzhu, baosiqi, tianxin06, hehuang, wang.fan, wu\_hua\}@baidu.com}}

\begin{document}

\maketitle

\begin{abstract}
In this paper, we propose an unsupervised query enhanced approach for knowledge-intensive conversations, namely QKConv. There are three modules in QKConv: a query generator, an off-the-shelf knowledge selector, and a response generator. QKConv is optimized through joint training, which produces the response by exploring multiple candidate queries and leveraging corresponding selected knowledge. The joint training solely relies on the dialogue context and target response, getting exempt from extra query annotations or knowledge provenances. To evaluate the effectiveness of the proposed QKConv, we conduct experiments on three representative knowledge-intensive conversation datasets: conversational question-answering, task-oriented dialogue, and knowledge-grounded conversation. Experimental results reveal that QKConv performs better than all unsupervised methods across three datasets and achieves competitive performance compared to supervised methods.
\end{abstract}

\section{Introduction}

In addition to open-domain chitchat, there exist various knowledge-intensive conversations, such as conversational question-answering, task-oriented dialogue, and knowledge-grounded conversation. Although large-scale language models can implicitly store common knowledge within parameters \citep{petroni2019language, zhao2020pre}, they are known to suffer from producing plausible statements with factual errors (a.k.a. knowledge hallucination) \citep{roller2020recipes, marcus2020next}. Therefore, there is a trend to rely on external resources, such as Wikipedia databases or search engine results, to facilitate knowledge-intensive conversations \citep{dinan2018wizard, komeili2022internet}.

In knowledge-intensive conversations, the most straightforward way to retrieve external knowledge is to take the dialogue context as the query and use an off-the-shelf retriever to return the knowledge entry. However, it encounters some difficulties in retrieving appropriate knowledge \citep{shuster2021retrieval}. As the focus or topic changes along with the conversation flow, the outdated information in the dialogue context brings extra noise to the retriever, resulting in obsolete or irrelevant knowledge retrieved. Moreover, the dialogue context has a native misalignment with the short and interrogative query preferred in existing retrievers.

Some methods choose to finetune a task-specific retriever to enhance the performance of knowledge selection \citep{guu2020retrieval, shuster2021retrieval, glass2022re2g}. However, this strategy is usually computationally expensive (e.g., finetuning a dense retriever requires constant recomputation for massive knowledge entries) or even infeasible for complex retrieval systems (e.g., retraining a search engine is impractical). Some other methods choose to generate a self-contained query based on the dialogue context \citep{yu2020few, anantha2021open, chen2022reinforced}. This strategy relies on careful query annotations to guarantee the completeness of essential information extraction and the adaptation to the knowledge selector.

\begin{figure*}[t]
    \centering
    \includegraphics[width=1\textwidth]{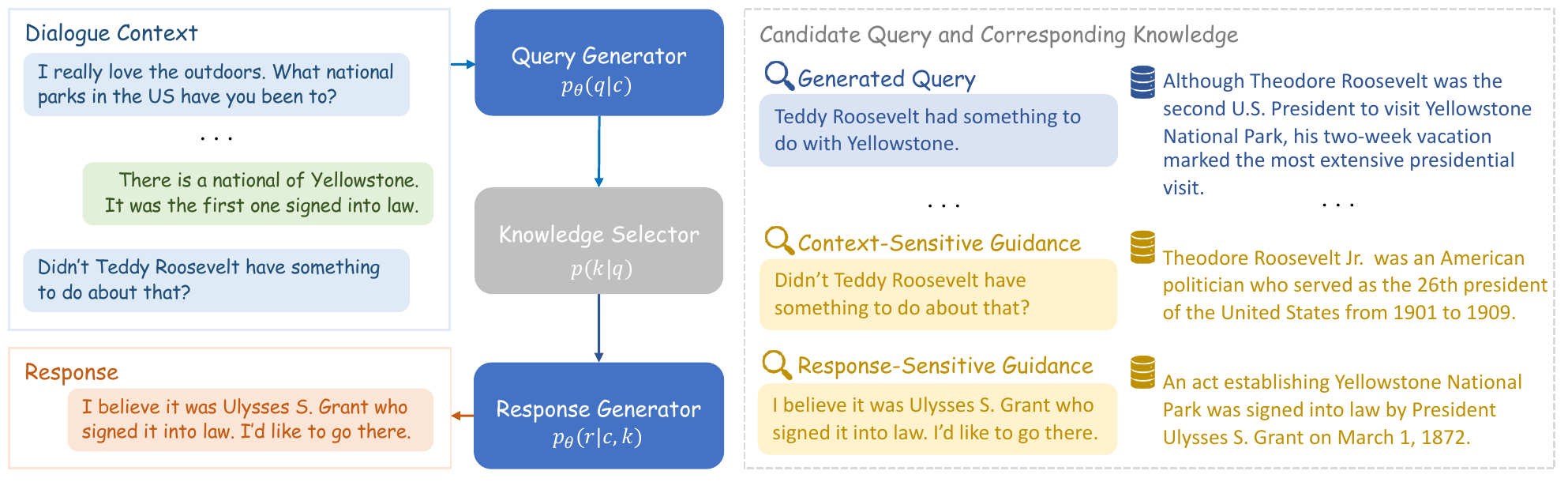}
    \caption{Overview of QKConv's joint training process. QKConv consists of three modules: a query generator, an off-the-shelf knowledge selector, and a response generator, where two generators share model parameters. During training, for a given dialogue context, QKConv learns to produce the target response by exploring multiple candidate queries and leveraging corresponding selected knowledge. Additionally, we integrate context-sensitive and response-sensitive guidance into the candidate query set to regulate query generation and facilitate joint training.}
    \label{fig:method}
\end{figure*}

In this paper, we introduce a novel unsupervised query enhanced approach for knowledge-intensive conversations, namely QKConv. As shown in Figure \ref{fig:method}, QKConv consists of three modules: a \textit{query generator}, an off-the-shelf \textit{knowledge selector}, and a \textit{response generator}. Specifically, QKConv is optimized through joint training, which produces the response by exploring multiple candidate queries and leveraging corresponding selected knowledge. We also integrate two types of query guidance to regulate query generation and facilitate joint training: \textit{context-sensitive} guidance (e.g., the last context utterance) and \textit{response-sensitive} guidance (e.g., the target response).

The benefits brought by QKConv's design are three-fold. Firstly, the training of QKConv solely relies on the dialogue context and target response, getting exempt from extra query annotations or knowledge provenances. Secondly, the joint training of QKConv boosts query generation toward better knowledge selection and ensures end-to-end performances, compared to the individual optimization of each module. Thirdly, thanks to the query generation module, QKConv gets rid of the expensive computation of tuning knowledge selectors and has the generality to adopt various knowledge selectors. 

To evaluate the effectiveness of the proposed QKConv, we conduct experiments on three representative knowledge-intensive conversation datasets: conversational question answering QReCC \citep{anantha2021open}, task-oriented dialogue SMD \citep{eric2017key}, and knowledge-grounded conversation WoW \citep{dinan2018wizard}. Experimental results reveal that QKConv performs better than all unsupervised methods across three datasets and even outperforms supervised methods on some datasets. 
Specifically, QKConv's generated query achieves superior knowledge selection performance, and QKConv exhibits robust knowledge utilization in response generation. 
We have released QKConv's code and model checkpoints\footnotemark[1], hoping to facilitate further research in knowledge-intensive conversations.
\footnotetext[1]{\url{https://github.com/PaddlePaddle/Knover/tree/develop/projects/QKConv}}

In summary, the main contributions of this paper are: (1) We propose an unsupervised query enhanced approach via joint training for knowledge-intensive conversations, namely {\M}. To the best of our knowledge, we are the first to utilize joint training for query generation. (2) We show that {\M} achieves state-of-the-art end-to-end results against all unsupervised methods and outperforms supervised methods on certain datasets. (3) We show that {\M} exhibits superior query quality and robust knowledge utilization in response generation.

\section{Methodology}
This paper introduces a query enhanced approach of QKConv, which incorporates query generation to boost knowledge-intensive conversations and optimizes the dialogue system via unsupervised joint training. As shown in Figure \ref{fig:method}, {\M} consists of three modules: \textit{Query Generator} to generate multiple queries based on the dialogue context; an off-the-shelf \textit{Knowledge Selector} to find relevant knowledge given queries; \textit{Response Generator} to produce the final response. In the following, we will elaborate the design of these modules and discuss the process of joint training in detail.

\subsection{Query Enhanced Knowledge-Intensive Conversation Modeling}

\subsubsection*{Query Generator}
The query generator aims to produce an effective query to retrieve appropriate knowledge for response generation. In the training process, with the dialogue context as input, the query generator will explore and produce multiple queries as candidates. The dialogue context is the concatenation of previous utterances $c=\{u_1, u_2, \dots, u_n\}$, and the candidate query $q\in \mathcal{Q}$ is generated with probability $p_\theta(q|c)$. 

\subsubsection*{Knowledge Selector}
The knowledge selector needs to find relevant knowledge from the knowledge base for a given query. To guarantee selection relevance, the off-the-shelf knowledge selector consists of one retriever for fast knowledge recall and one successive reranker for fine-grained relevance estimation. Given a candidate query $q$, the final knowledge selection score is the combination of two-stage scores \cite{gallagher2019joint}: 
\begin{equation}
    p(k|q) = \sigma\big(S_{retrieval}(k|q) + S_{rerank}(k|q)\big)
\end{equation}
where $\sigma(\cdot)$ refers to the sigmoid function. Unless specified, the knowledge with the highest score will be selected for the given query and used in the response generation.

\subsubsection*{Response Generator}
The response generator aims to produce an appropriate response grounded on selected knowledge. In the training process, with the dialogue context and candidate knowledge as input, the probability of producing the target response is estimated as $p_\theta(r|c, k)$. In addition, the response and query generators share model parameters, with prompts added for task differentiation\footnotemark[2].
\footnotetext[2]{The prompt used in the query generator is "\texttt{translate dialogue context to query:}". The prompt used in the response generator is "\texttt{generate system response based on knowledge and dialogue context:}".}

\subsection{Joint Training}

\label{sec:JT}
Under such a design, the response generation in knowledge-intensive conversations is modeled as follows:
\begin{equation}
    p(r|c) \propto \sum_{q\in \mathcal{Q}} p_\theta(q|c)~p(k|q)~p_\theta(r|c,k)
\end{equation}
where $c$ is the dialogue context, $r$ is the target response, $q$ is one candidate query, and $k$ is its corresponding knowledge. The training objective is to maximize the generation probability of the target response through marginalization over candidate queries. Exploring multiple query candidates leads to diverse knowledge selection and generation probability of target response. Supposing one candidate query stimulates the knowledge coherent with the dialogue context and relevant to the target response, the joint training will encourage this query generation and facilitate knowledge utilization in response generation. Otherwise, the joint optimization will suppress the corresponding query generation and restrain knowledge utilization in response generation.
 
During training, we propose to integrate \textit{context-sensitive} guidance (e.g., the last context utterance $u_n$) and \textit{response-sensitive} guidance (e.g., the target response $r$) into the candidate query set. The benefits brought by the guidance integration are two-fold. Firstly, the query guidance can regulate query generation. Context-sensitive guidance suggests extracting essential information from the context, and response-sensitive guidance suggests predicting the focus of the target response. These two guidance act as references and help the query generator avoid non-sense queries in unsupervised training. Secondly, the two types of query guidance can facilitate joint training. Since selecting the relevant knowledge for the target response is challenging, constant exposure to irrelevant knowledge will make the model ignore given knowledge and generate generic responses. Incorporating context-sensitive (prior) and response-sensitive (posterior) guidance amplifies knowledge diversity and enhances the selection of relevant knowledge. The exposure to diverse knowledge (relevant and irrelevant) helps facilitate end-to-end joint training. In short, such incorporation helps avoid the degradation of non-sense query generation and knowledge-independent response generation in joint training.

To alleviate the costly query generation and knowledge selection at each training step, we utilize iterative training to speed up the training process, which embraces an inner-outer loop structure for model training and data collection. In the outer loop, the inference is carried out over the train set to collect candidate queries with the up-to-date query generator and corresponding knowledge with the off-the-shelf knowledge selector. In the inner loop, the query and response generators are optimized jointly to maximize the probability of the target response. The inner-outer loop will iterate several times until convergence.

\section{Experiments}

\subsection{Experiment Settings}

\subsubsection{Datasets}
We conduct experiments on three datasets over diverse knowledge-intensive conversation tasks: QReCC \cite{anantha2021open} for conversational question answering, Standford Multi-Domain (SMD) \cite{eric2017key} for task-oriented dialogue, and Wizard of Wikipedia (WoW) \cite{dinan2018wizard} for open-domain knowledge-grounded dialogue.

QReCC\footnotemark[3] contains 14K open-domain conversations with 80K question-answer pairs, where each conversational question is rewritten into a self-contained query by human crowdworkers. The knowledge base is a collection of 54M passages split from 10M web pages and indexed by BM25.
\footnotetext[3]{The version of QReCC dataset is \url{https://zenodo.org/record/5115890}. We remove conversations without truth responses. The validation set without an official version is randomly selected 5\% from the training set.}

SMD is a task-oriented dialogue dataset including 3K conversations. Each conversation is equipped with a small knowledge base.

Wizard of Wikipedia (WoW)\footnotemark[4] is an open-domain dialogue dataset with 18K conversations. The conversations are grounded on knowledge from Wikipedia retrieved by TF-IDF.
\footnotetext[4]{We use the version of WoW dataset in the KILT benchmark \cite{petroni2021kilt}. The knowledge source is a collection of 5.9M Wikipedia pages.}

\subsubsection{Baselines}

We compare QKConv to the previous state-of-the-art supervised and unsupervised models on each dataset. 
Details about the compared models are summarized in Table \ref{tab:sota}. 
Supervised models leverage either query annotations or knowledge selection annotations, while unsupervised models only rely on the dialogue context and response. 
Among these models, tuning dense retrievers is employed in DPR (IHN)-FiD \cite{kim2022saving}, Re2G \cite{glass2022re2g}, Hindsight \cite{paranjape2022hindsight}, while the query generation method is preferred by Q-TOD \cite{tian2022q} and \citet{raposo2022question}.
Compared to methods augmented by knowledge selection, UnifiedSKG \cite{xie2022unifiedskg} utilizes the entire knowledge base to generate the response.

\begin{table*}[ht]
\setlength\tabcolsep{5pt}
\begin{center}
\small
\renewcommand{\arraystretch}{1.1}
\begin{tabular}{@{}ll p{0.3\textwidth} p{0.18\textwidth} p{0.15\textwidth} @{}}
\toprule
Datasets & Metrics & Compared Model & Extra Supervision & Pre-trained Model \\ 
\midrule
\multirow{2}{*}{QReCC} &   \multirow{2}{*}{F1, EM} & DPR(IHN)-FiD \cite{kim2022saving}$^\dagger$     & Selection Annotations            & T5-base         \\
& & \citet{raposo2022question}$^\ddagger$     & -  & pegasus-large\\
\multirow{2}{*}{SMD}   &   \multirow{2}{*}{Entity-F1, BLEU} & Q-TOD \cite{tian2022q}$^\dagger$    & Query Annotations   & T5-large \\
& & UnifiedSKG \cite{xie2022unifiedskg}$^\ddagger$    & -   & T5-large \\
\multirow{2}{*}{WoW}   &   \multirow{2}{*}{KILT-F1, KILT-Rouge-L} & Re2G \cite{glass2022re2g}$^\dagger$ & Selection Annotations   & BART-large \\
& & Hindsight \cite{paranjape2022hindsight}$^\ddagger$ & -   & BART-large \\
\bottomrule
\end{tabular}
\end{center}
\caption{Summarization of the state-of-the-art supervised and unsupervised models. $^\dagger$ and $^\ddagger$ denote the state-of-the-art supervised and unsupervised models, respectively.}
\label{tab:sota}
\end{table*}

\begin{table*}
\begin{center}
\small
\renewcommand{\arraystretch}{1.1}
\begin{tabular}{@{}p{0.21\textwidth} p{0.1\textwidth}<{\centering} p{0.1\textwidth}<{\centering} p{0.1\textwidth}<{\centering} p{0.1\textwidth}<{\centering} p{0.1\textwidth}<{\centering} p{0.1\textwidth}<{\centering}}
\toprule
\multirow{2}{*}{}     & \multicolumn{2}{c}{QReCC}      & \multicolumn{2}{c}{SMD}         & \multicolumn{2}{c}{WoW}         \\
                      & F1             & EM            & Entity F1      & BLEU           & KILT-F1             & KILT-RL        \\ \midrule
Previous SOTA (w/ label)         & 30.40           & 4.70           & \underline{71.11}          & \underline{21.33}          & 12.98          & 11.39          \\
Previous SOTA (w/o label)        & 18.90           & 1.00           & 65.85          & 17.27          & 13.39          & 11.92          \\ 
\M        & \underline{\textbf{33.54}} & \underline{\textbf{5.90}}          & \textbf{68.94}          & \textbf{20.35} & \underline{\textbf{13.64}} & \underline{\textbf{12.03}} \\ \bottomrule
\end{tabular}
\end{center}
\caption{Evaluation results on SMD, QReCC, and WoW test sets, with the best value of the dataset indicated by underlines and the best value from unsupervised methods written in bold.}
\label{tab:main}
\end{table*}

\subsubsection{Implementation Details}
\start{Knowledge Selector}
Following the retriever setting of the original dataset, BM25 and TF-IDF are employed for QReCC and WoW, respectively.
However, the SMD dataset does not involve a retriever due to the fairly small knowledge base. 
For reranking, an off-the-shelf model RocketQA \cite{ren2021rocketqav2} is used for all datasets. 

\start{Generator}
We employ the same pre-trained model as the state-of-the-art supervised model to perform query and response generation, i.e., T5-base (220M) \cite{raffel2020exploring} for QReCC, T5-large (770M) \cite{raffel2020exploring} for SMD, and BART-large (400M) \cite{lewis2020bart} for WoW.

\start{Training}
{\M} is trained in an inner-outer loop structure that iteratively executes query generation, knowledge selection in the outer loop, and model updating in the inner loop. For query generation, we adopt beam search with a beam size of 4 as the decoding strategy and use all decoding results as candidate queries. Therefore, the set of query candidates consists of four generated queries, one response-sensitive guidance, and one context-sensitive guidance. The response-sensitive guidance refers to the target response. In light of previous common queries \cite{raposo2022question, shuster2021retrieval}, the context-sensitive guidance refers to the last utterance of dialogue on QReCC and dialogue context on SMD and WoW. To familiarize pre-trained models with dialogue tasks, the generator is warmed up with the response generation task for a few epochs. 

\start{Inference}
The decoding strategy of query and response generation is beam search with a beam size of 4. We use the decoding result with the highest probability as the final result.

More details about hyperparameter settings are provided in Appendix \ref{sec:appendixHyper}.

\subsection{Results}
% summary
We evaluate the end-to-end performance of our models on the three knowledge-intensive dialogue datasets following the metrics used in prior studies \citep{anantha2021open, eric2017key, petroni2021kilt}. 
In particular, Entity-F1 \cite{eric2017key} measures overlapping entities between generated response and ground truth. KILT-F1 and KILT-Rouge-L (KILT-RL) \cite{petroni2021kilt} only award points to instances with accurate knowledge selection.
Table \ref{tab:main} summarizes the results of our models and the state-of-the-art models trained with and without supervision on three datasets.

QKConv consistently outperforms the unsupervised results on three datasets and even surpasses the supervised results on QReCC and WoW.
Compared to unsupervised models, on the F1 score, QKConv achieves a relative improvement of 78.2\% on QReCC, 4.7\% on SMD, and 1.9\% on WoW, respectively. 
The encouraging improvements demonstrate that our proposed QKConv has strong effectiveness and robustness to generate high-quality responses across various knowledge-intensive conversations. 
In comparison to supervised SOTA with retriever finetuning, QKConv obtains the best F1 scores with a relative increment of 10.8\% on QReCC, and 5.1\% on WoW, respectively.
As for the supervised models with query annotations, the relatively lower Entity-F1 on SMD suggests some room for improvement for unsupervised QKConv.

\section{Discussion}

In this section, to further dissect the proposed QKConv, more experiments are conducted on the QReCC dataset. Unless specified, the pre-trained model of T5-large is employed in the following experiments. 

\subsection{Query Generation Analysis}
\label{sec:queryAna}
In this paper, a query enhanced approach is introduced for knowledge-intensive conversations. For an in-depth analysis of query incorporation, we will discuss three research questions regarding QKConv's query on essential, modality, and superiority.

\paragraph{RQ$_1$} Is it \textbf{\textit{essential}} to generate queries for knowledge selection?

It is known that the most straightforward way is to employ the dialogue context or the last utterance as the query for knowledge selection. We compare the impact of various query types on knowledge selection, with results summarized in Table \ref{tab:query}.\footnotemark[5] The knowledge selection results by the target response and golden query are also provided for reference. Measure by the Recall@1 score, QKConv's generated query improves knowledge selection performance by 4.16\% compared to the dialogue context and narrows the gap to 5.75\% compared to the golden query. In addition, the improvement reaches 34.04\% compared to the widely adopted last utterance. These results suggest that query generation is essential in boosting knowledge selection. 
\footnotetext[5]{Following \citet{wu2021conqrr,kim2022saving}, instances without ground truth are ignored in evaluating knowledge selection.}

\begin{table}[t]
\begin{center}
\small
\renewcommand{\arraystretch}{1.1}
\begin{tabular}{@{}l p{0.1\textwidth}<{\centering} ccc@{}}
\toprule
\multirow{2}{*}{Query}  & Knowledge   &   \multicolumn{3}{c}{Query Statistics} \\
&    Recall@1   &   Length  &   C-F1  &   R-F1        \\ \midrule
Context &   39.15   &   89.55   &   100 &   15.54    \\
Last Utterance  &   9.27    &   6.44    &   29.95   &   11.83   \\ 
Response  &   83.32   &   19.34   &   15.54   &   100   \\
Golden Query    &   49.06   & 9.89  &   33.10   &   23.93   \\ \cmidrule{1-5}
{\M}   &   43.31   &19.49  &   48.01   &   23.05   \\\bottomrule
\end{tabular}
\end{center}
\caption{Knowledge selection results and corresponding query statistics on the QReCC test set. C-F1 and R-F1 are abbreviated for Context-F1 and Response-F1.}
\label{tab:query}
\end{table}

\paragraph{RQ$_2$} What is the generated query's \textbf{\textit{modality}}, similar to the dialogue context or the response?

As described in Section \ref{sec:JT}, QKConv incorporates context-sensitive and response-sensitive guidance to regulate query generation. After joint training, what is the modality of the generated query, similar to the dialogue context or the response? For this investigation, we estimate the similarity of the generated query to the dialogue context and the target response using the word overlapping F1 metric. The Context-F1 and Response-F1 results are summarized in Table \ref{tab:query}, together with the query length statistics. 

The relatively high value of Context-F1 indicates that the generated query gathers intensive information from the context. Meanwhile, the relatively high value of Response-F1 indicates that the generated query includes relevant information with the response. In short, the generated query exhibits a hybrid modality, incorporating intensive information from the dialogue context and some predicted hints toward the response. One qualitative example is also provided in Table \ref{tab:queryCase} to illustrate this phenomenon.  

\paragraph{RQ$_3$} Is the performance of the generated query \textbf{\textit{superior}} to other state-of-the-art approaches?

On the QReCC dataset, CONQRR \citep{wu2021conqrr} is the state-of-the-art query generation approach, which leverages query annotations and a reward function to optimize the query generation through supervised and reinforcement learning. CONQRR utilizes the BM25 retriever as the knowledge selector and employs T5-base as the pre-trained model. Table 4 summarizes the knowledge selection performance of CONQRR and QKConv. 

When compared under the same retriever, despite that QKConv is optimized via unsupervised joint training, the generated query achieves 4.79\% higher MRR@10 than CONQRR. 
The remarkable improvement of generated queries confirms the superior performance of {\M} on knowledge selection.
In addition, QKConv equipped with a reranker raises MRR@10 by 6.52\% and Recall@1 by 5.39\% significantly. These results confirm the benefits of adopting the combinatorial knowledge selector.

\begin{table}[t]
\setlength\tabcolsep{3pt}
\begin{center}
\small
\renewcommand{\arraystretch}{1.1}
\begin{tabular}{@{}llcc@{}}
\toprule
Model  &  Knowledge Selector   &   MRR@10 &  Recall@1       \\ \midrule
% Golden Query & Retriever & 39.80 & 27.02 \\
CONQRR &   Retriever   &   38.30   & -     \\ \cmidrule{1-4}
\multirow{2}{*}{\M}  &   Retriever    &   43.09    & 36.34    \\ 
 &   Retriever+Reranker  &   49.61  &  41.73   \\ 
 \bottomrule
\end{tabular}
\end{center}
\caption{Comparison of knowledge selection performance between QKConv and CONQRR with T5-base as the pre-trained model.}
\label{tab:queryQuality}
\end{table}

\subsection{Knowledge Utilization Ability}

{\M} also demonstrates strong knowledge utilization ability in response generation, apart from accurate knowledge selection in query generation.
As the selected knowledge is not always appropriate, the response generator encounters the challenge of properly utilizing the selected knowledge. When confronting appropriate knowledge, the response generator is expected to ground on the knowledge and then incorporate it properly. In contrast, with irrelevant knowledge, the response generator should denoise and eliminate high reliance on it. 

To investigate the knowledge utilization ability of {\M}, we divide the selected knowledge into accurate and inaccurate knowledge according to the Recall@1 metrics. We compare the response generator of {\M} with the response generator baseline. The baseline model is trained in an individually optimized manner (not joint training), with the dialogue context and knowledge selected by golden queries as input and the target response as output. In the evaluation phase, the same data is applied for comparisons.

\paragraph{Automatics evaluation}
We compute the F1 score between generated responses and ground truth and the KR-F1 score for both models. The KR-F1 score, adapted from \citet{paranjape2022hindsight}, evaluates the F1 score between generated response and selected knowledge (not golden knowledge). The optimal value for KR-F1 is the one \textit{being close to the KR-F1 by ground truth}, which indicates a natural knowledge utilization rather than under-utilization or over-reliance. 

Table \ref{tab:KU} summarizes knowledge utilization ability with ground-truth results as references. 
For the overall F1 score, {\M} outperforms the baseline model by 1.87\%.
Considering results based on knowledge correctness, the KR-F1 for correct knowledge is more significant than incorrect knowledge by 3.73\% in {\M}. The notable gap reveals that {\M} can distinguish knowledge associated with dialogue context and rely more on the correct knowledge. A similar but smaller gap (2.13\%) can be found in the baseline model, which suggests that this ability is introduced by exposing diverse knowledge quality to response generation during training. Furthermore, with the correct knowledge, {\M} demonstrates a significantly higher F1 and closer KR-F1 than the baseline model. 

\begin{table}[t]
\setlength\tabcolsep{5pt}
\begin{center}
\small
\renewcommand{\arraystretch}{1.1}
\begin{tabular}{@{}lccccc@{}}
\toprule
\multirow{2}{*}{Model} & Overall  & \multicolumn{2}{c}{Recall@1=1}   &   \multicolumn{2}{c}{Recall@1=0} \\
& F1 & F1 & KR-F1 & F1 & KR-F1 \\
 \midrule
Baseline   &  34.40  & 60.98 & 15.29 & 21.61 & 13.16 \\
{\M} & \textbf{36.27} & \textbf{63.20} &   \textbf{14.31} &  \textbf{23.55} & \textbf{10.58} \\ \cmidrule{1-6}
Ground Truth & 100 & 100 & 12.72 & 100 & 6.18 \\
\bottomrule
\end{tabular}
\end{center}
\caption{Comparisons of knowledge utilization ability between {\M} and individually optimized baseline model, with the best value written in bold. KR-F1 evaluates the overlap between generated response and selected knowledge.}
\label{tab:KU}
\end{table}

\paragraph{Human evaluation} 
We randomly sampled 50 examples with correct knowledge and another 50 with incorrect knowledge. Crowd-sourcing workers evaluate each sample on three aspects with a range of [0, 1, 2]:

\begin{itemize}[leftmargin=*,noitemsep,topsep=0pt]
    \item Coherence assesses whether the response is relevant and consistent with the dialogue context.
    \item Groundedness assesses whether the response contains information from the given knowledge.
    \item Engagingness  measures the willingness to have a long conversation.
\end{itemize}

Table \ref{tab:HE} demonstrates that {\M} outperforms the baseline model regarding Coherence and Engagingness, while achieving similar levels of Groundedness with accurate knowledge and lower Groundedness (by 0.27) with inaccurate knowledge. These results indicate that compared to the individually-optimized baseline, {\M} can incorporate correct knowledge to a more natural degree and yield higher-quality responses. 

In short, both automatic and human evaluation results confirm that {\M} attains robustness to different qualities of knowledge and a remarkable knowledge utilization ability to correct knowledge.

\begin{table}[t]
\setlength\tabcolsep{5pt}
\begin{center}
\small
\renewcommand{\arraystretch}{1.1}
\begin{tabular}{@{}lccc@{}}
\toprule
Model & Coherence  &  Groundedness  &   Engagingness \\
 \midrule
&  \multicolumn{3}{c}{Recall@1=1} \\ \cmidrule{2-4}
Baseline   &  1.64  & 2 & 1.63 \\
{\M} & \textbf{1.78} & 2 & \textbf{1.76} \\ \cmidrule{1-4}
&  \multicolumn{3}{c}{Recall@1=0} \\ \cmidrule{2-4}
Baseline   &  0.89  & \textbf{1.87} & 0.84 \\ 
{\M} & \textbf{1.16} & 1.60 & \textbf{1.11} \\
\bottomrule
\end{tabular}
\end{center}
\caption{Human evaluation results with the best scores written in bold.}
\label{tab:HE}
\end{table}

\subsection{Effect of Guidance}

% 动机、方法
We propose context-sensitive and response-sensitive guidance to regulate query generation and facilitate joint training. The query generation demonstrates a hybrid modality under the regulation of guidance as described in Section \ref{sec:queryAna}. To scrutinize the efficacy of guidance in joint training, we conduct ablation experiments with {\M}, detailed in Table \ref{tab:guidance}.

% 结果、结论
In the absence of all guidance, our model witnesses a marked decrease in all metrics, resulting in 2.92\%/1.09\%/2.93\% declines in F1/EM/Recall@1. With the incorporation of either guidance, knowledge selection and end-to-end performances are enhanced to a considerable extent but remain inferior to {\M}. These results indicate that both types of guidance contribute to joint training, and the combined implementation yields the most significant benefits. 
Despite the decline in performance, {\M} trained without guidance still outperforms the state-of-the-art models (\citet{raposo2022question} with 18.90 F1 and 1.00 EM), highlighting that the advantages of our method are brought by joint training and boosted by two types of query guidance.

\begin{table}[t]
\setlength\tabcolsep{3pt}
\begin{center}
\small
\renewcommand{\arraystretch}{1.1}
\begin{tabular}{@{}l p{0.08\textwidth}<{\centering} p{0.08\textwidth}<{\centering} p{0.082\textwidth}<{\centering}@{}}
\toprule
  & F1   &   EM &  Recall@1       \\ \midrule
{\M} &   36.27   &   7.03   & 43.31     \\ \cmidrule{1-4}
\quad no guidance  &   33.35$_{\downarrow 2.92}$    &   5.94$_{\downarrow 1.09}$    & 40.38$_{\downarrow 2.93}$       \\ 
\quad w/ context-sensitive  &   35.24$_{\downarrow 1.03}$    &   6.35$_{\downarrow 0.68}$    & 42.76$_{\downarrow 0.55}$       \\ 
\quad w/ response-sensitive  &   34.75$_{\downarrow 1.52}$    &   6.46$_{\downarrow 0.57}$    & 41.97$_{\downarrow 1.34}$       \\ 
\bottomrule
\end{tabular}
\end{center}
\caption{Comparisons of the effect of guidance.}
\label{tab:guidance}
\end{table}

\subsection{Case Studies}
\label{sec:case}

\begin{table*}[t]
\setlength\tabcolsep{3pt}
\begin{center}
\small
\renewcommand{\arraystretch}{1.1}
\begin{tabular}{@{}l p{0.8\textwidth}@{}}
\toprule
\multicolumn{2}{@{}l}{\textbf{Cherry-picked example}} \\ \midrule
Dialogue History & \makecell[tl]{\textbf{User}: what were some of john stossel's most popular publications? \\ 
\textbf{System}: \textcolor{my_blue}{\textit{give me a break: how i exposed hucksters, cheats, and scam artists and became the scourge}} \\ \textcolor{my_blue}{\textit{of the liberal media}} is an autobiography documenting stossel's career and philosophical transition.\\
...  \\
\textbf{User}: what was the response?} \\
{\M} Query  & give me a break: how i exposed hucksters, cheats, and scam artists and became the scourge of the liberal media received generally positive reviews from critics. (\usym{1F5F8})\\
Selected Knowledge & give me a break: how I ... \underline{it was a new york times bestseller for 11 weeks.} ... \\
{\M} Response &  it was a new york times bestseller for 11 weeks. \\
\midrule
\multicolumn{2}{@{}l}{\textbf{Lemon-picked example}} \\ \midrule
Dialogue History & \makecell[tl]{\textbf{User}: what part did victor mclaglen play in happy days?  \\ 
\textbf{System}:  \textcolor{my_blue}{\textit{victor mclaglen was a minstrel show performer in the film, happy days.}} \\
...  \\
\textbf{User}: what other films did he play in?} \\
{\M} Query  & victor mclaglen was a minstrel show performer in the film, happy days. (\usym{2717}) \\
Selected Knowledge & originally titled \underline{new orleans frolic}, the story centers around margie (played by marjorie white ), ... \underline{victor mclaglen as minstrel show performer} ... \\
{\M} Response & victor mclaglen played a minstrel show performer in the film, new orleans frolic. \\
\bottomrule
\end{tabular}
\end{center}
\caption{Examples of queries generated by {\M} on QReCC test set. \textcolor{my_blue}{\textit{Blue}} marks the provenance of queries, and the underline highlights the source of response. \usym{1F5F8}/\usym{2717} inside the bracket indicates top-1 knowledge selection accuracy.}
\label{tab:queryCase}
\end{table*}

We provide a cherry-picked example and a lemon-picked example in Table \ref{tab:queryCase} to gain insight into the performance of {\M}. Additional examples are available in Appendix \ref{sec:appendixCase}.

The cherry-picked example inquires about the reaction of a previously stated book. For query generation, the query generated by {\M} is response-looking, attempting to reply to the conversation. Although the response-looking query contains certain counterfeit information, the book's full title extracted from the conversation history contributes to accurate knowledge selection. For response generation, {\M} locates the relevant sentence within the long knowledge paragraph and generates an appropriate response. 

The lemon-picked example inquires about an actor's films in addition to the previously mentioned one. Our model's generated query is also response-looking, extracting relevant information from the previous text and organizing it into a response. However, the model fails to consider the limiting word "other" in the last utterance, resulting in inappropriate knowledge selection and a similar response as in the previous dialogue history.

\section{Related Work}

\paragraph{Knowledge-Intensive Conversation}
To attend knowledge in conversations, some prior studies concentrate on how to ground the given knowledge \cite{ma2020compare, xie2022unifiedskg} or elicit parametric knowledge from large language models \cite{zhao2020pre}. Recently, access to an external knowledge corpus has attracted a spate of interest, in line with our paper, and has come up with several datasets. For instance, some datasets provide a fixed small knowledge base for each sample \citep{eric2017key, wen2017network, dinan2018wizard, moghe2018towards}. In more natural circumstances, using a uniform large-scale knowledge base for all samples, such as Wikipedia dumps, web crawl data, or even search engines, has become a trend \citep{zhou2018dataset, petroni2021kilt, anantha2021open, komeili2022internet}. However, it should be noted that knowledge selection challenges increase with the size of the knowledge base, and selection performance bounds the performance of response generation. Therefore, the performance of knowledge selection is crucial for knowledge-intensive dialogue. Two primary directions to address knowledge selection are finetuning knowledge selectors and generating a context-independent query.

\paragraph{Retrieval-Augmented Generation}
Recently, an increasing interest has been shown in modeling a dense knowledge selector and response generator simultaneously, with the dialogue context as the query. Many of these works utilize joint training \citep{lewis2020retrieval, guu2020retrieval, shuster2021retrieval, huang2021plato, thulke2021efficient, glass2022re2g} or reinforcement learning \citep{zhao2020knowledge} to modify the prior selection distribution. As opposed, some studies directly involve the posterior distribution of knowledge to enhance knowledge selection \citep{lian2019learning, Kim2020Sequential, paranjape2022hindsight}. However, repeated index rebuilding for the updated knowledge selector is time-consuming with the large-scale knowledge base, and the involvement of posterior distribution may render the training-inference discrepancy. Furthermore, a few works consider a complicated selection process attributed to the challenging and interrupted gradient propagation \cite{glass2022re2g}. This paper investigates the query generator rather than the selector and exploits off-the-shelf selectors to refrain from the above problems.

\paragraph{Query Generation}
A lengthy dialog context as the query reduces the efficiency of the knowledge selector and may be misaligned with the form preferred in off-the-shelf selectors. Prior works \citep{yu2020few, anantha2021open, vakulenko2021question, komeili2022internet, tian2022q} leverage query annotations as supervision to train query generators that convert a dialogue context into a context-independent query, but facing the problem of human-written queries often unavailable in practice.  With the absence of external supervision, \citet{mao2021generation} regards response and knowledge as training targets to expand the original query. However, memorizing response and knowledge has a heavy burden on the model for a large-scale knowledge base. Moreover, some current studies argue that the supervised learning of queries disconnects from knowledge selection and end-to-end performance \citep{wu2021conqrr, chen2022reinforced}. Instead, they exploit reinforcement learning with extra query and retrieval annotations to generate queries adaptive to downstream performance. In this paper, we propose a novel query enhanced approach that jointly trains the query generator with the response generator without additional supervision. The end-to-end training also ensures the downstream performance of queries. Furthermore, our approach with two query guidance gets exempt from the risk of generating unreadable sentences experienced frequently in reinforcement learning \cite{ouyang2022training}.

\section{Conclusion}

This paper introduces a query enhanced approach of QKConv for knowledge-intensive conversations, which is optimized via unsupervised joint training without any reliance on query annotations or knowledge provenances. The experiments are carried out on three knowledge-intensive conversation datasets: conversational question answering QReCC, task-oriented dialogue SMD, and knowledge-grounded conversation WoW. The proposed QKConv outperforms all unsupervised methods across three datasets. Compared to supervised methods, QKConv even establishes new state-of-the-art results on QReCC and WoW. Further analysis demonstrates that with joint training, the query generation adapts well to the knowledge selector, and the response generation has utilization robustness towards various knowledge.

\section*{Limitations}
As shown in Table \ref{tab:main}, our approach underperforms the state-of-the-art supervised model on the SMD dataset, where the supervised SOTA labels a search instruction for each sample. In addition, the lemon-picked example in Table \ref{tab:queryCase} demonstrates that sometimes it is challenging for the query generator to learn complicated expressions automatically. Despite our model's superiority over all unsupervised methods, these gaps reveal some improvement room of {\M}. In Appendix \ref{sec:fewshot}, we try to bridge the gaps by incorporating a few query annotations.
Another limitation is that our approach suffers from the time-consuming off-the-shelf knowledge selection when given a large dataset and knowledge base. It takes half of the training hours in knowledge selection since it involves heavy computation of retrieval from a large-scale knowledge base and reranking with a cross-encoder.

\section*{Acknowledgement}
We would like to thank the anonymous reviewers for valuable comments. We thank Hua Lu and Yingzhan Lin for helpful discussions; Jingzhou He, Shiwei Huang, and Dou Hong for the help on resource coordination.

\bibliography{bibtex}

\begin{thebibliography}{37}
\expandafter\ifx\csname natexlab\endcsname\relax\def\natexlab#1{#1}\fi

\bibitem[{Anantha et~al.(2021)Anantha, Vakulenko, Tu, Longpre, Pulman, and
  Chappidi}]{anantha2021open}
Raviteja Anantha, Svitlana Vakulenko, Zhucheng Tu, Shayne Longpre, Stephen
  Pulman, and Srinivas Chappidi. 2021.
\newblock \href {https://aclanthology.org/2021.naacl-main.44} {Open-domain
  question answering goes conversational via question rewriting}.
\newblock In \emph{Proceedings of the 2021 Conference of the North American
  Chapter of the Association for Computational Linguistics: Human Language
  Technologies}, pages 520--534.

\bibitem[{Chen et~al.(2022)Chen, Zhao, Fang, Fetahu, Rokhlenko, and
  Malmasi}]{chen2022reinforced}
Zhiyu Chen, Jie Zhao, Anjie Fang, Besnik Fetahu, Oleg Rokhlenko, and Shervin
  Malmasi. 2022.
\newblock \href {"https://arxiv.org/abs/2210.15777"} {Reinforced question
  rewriting for conversational question answering}.
\newblock \emph{arXiv preprint arXiv:2210.15777}.

\bibitem[{Dinan et~al.(2019)Dinan, Roller, Shuster, Fan, Auli, and
  Weston}]{dinan2018wizard}
Emily Dinan, Stephen Roller, Kurt Shuster, Angela Fan, Michael Auli, and Jason
  Weston. 2019.
\newblock \href {https://openreview.net/forum?id=r1l73iRqKm} {Wizard of
  wikipedia: Knowledge-powered conversational agents}.
\newblock In \emph{International Conference on Learning Representations}.

\bibitem[{Eric et~al.(2017)Eric, Krishnan, Charette, and Manning}]{eric2017key}
Mihail Eric, Lakshmi Krishnan, Francois Charette, and Christopher~D Manning.
  2017.
\newblock \href {"https://aclanthology.org/W17-5506"} {Key-value retrieval
  networks for task-oriented dialogue}.
\newblock In \emph{Proceedings of the 18th Annual SIGdial Meeting on Discourse
  and Dialogue}, pages 37--49.

\bibitem[{Gallagher et~al.(2019)Gallagher, Chen, Blanco, and
  Culpepper}]{gallagher2019joint}
Luke Gallagher, Ruey-Cheng Chen, Roi Blanco, and J~Shane Culpepper. 2019.
\newblock \href {https://doi.org/10.1145/3289600.3290986} {Joint optimization
  of cascade ranking models}.
\newblock In \emph{Proceedings of the twelfth ACM international conference on
  web search and data mining}, pages 15--23.

\bibitem[{Glass et~al.(2022)Glass, Rossiello, Chowdhury, Naik, Cai, and
  Gliozzo}]{glass2022re2g}
Michael Glass, Gaetano Rossiello, Md~Faisal~Mahbub Chowdhury, Ankita Naik,
  Pengshan Cai, and Alfio Gliozzo. 2022.
\newblock \href {https://aclanthology.org/2022.naacl-main.194} {{R}e2{G}:
  Retrieve, rerank, generate}.
\newblock In \emph{Proceedings of the 2022 Conference of the North American
  Chapter of the Association for Computational Linguistics: Human Language
  Technologies}, pages 2701--2715, Seattle, United States. Association for
  Computational Linguistics.

\bibitem[{Guu et~al.(2020)Guu, Lee, Tung, Pasupat, and
  Chang}]{guu2020retrieval}
Kelvin Guu, Kenton Lee, Zora Tung, Panupong Pasupat, and Mingwei Chang. 2020.
\newblock \href {https://proceedings.mlr.press/v119/guu20a.html} {Retrieval
  augmented language model pre-training}.
\newblock In \emph{International Conference on Machine Learning}, pages
  3929--3938. PMLR.

\bibitem[{Huang et~al.(2021)Huang, He, Bao, Wang, Wu, and
  Wang}]{huang2021plato}
Xinxian Huang, Huang He, Siqi Bao, Fan Wang, Hua Wu, and Haifeng Wang. 2021.
\newblock \href {"https://aclanthology.org/2021.nlp4convai-1.14"} {Plato-kag:
  Unsupervised knowledge-grounded conversation via joint modeling}.
\newblock In \emph{Proceedings of the 3rd Workshop on Natural Language
  Processing for Conversational AI}, pages 143--154.

\bibitem[{Kim et~al.(2020)Kim, Ahn, and Kim}]{Kim2020Sequential}
Byeongchang Kim, Jaewoo Ahn, and Gunhee Kim. 2020.
\newblock \href {https://openreview.net/forum?id=Hke0K1HKwr} {Sequential latent
  knowledge selection for knowledge-grounded dialogue}.
\newblock In \emph{International Conference on Learning Representations}.

\bibitem[{Kim and Kim(2022)}]{kim2022saving}
Sungdong Kim and Gangwoo Kim. 2022.
\newblock \href {http://arxiv.org/abs/2202.07280v1} {Saving dense retriever
  from shortcut dependency in conversational search}.
\newblock \emph{arXiv preprint arXiv:2202.07280v1}.

\bibitem[{Komeili et~al.(2022)Komeili, Shuster, and
  Weston}]{komeili2022internet}
Mojtaba Komeili, Kurt Shuster, and Jason Weston. 2022.
\newblock \href {"https://aclanthology.org/2022.acl-long.579"}
  {Internet-augmented dialogue generation}.
\newblock In \emph{Proceedings of the 60th Annual Meeting of the Association
  for Computational Linguistics (Volume 1: Long Papers)}, pages 8460--8478.

\bibitem[{Lewis et~al.(2020{\natexlab{a}})Lewis, Liu, Goyal, Ghazvininejad,
  Mohamed, Levy, Stoyanov, and Zettlemoyer}]{lewis2020bart}
Mike Lewis, Yinhan Liu, Naman Goyal, Marjan Ghazvininejad, Abdelrahman Mohamed,
  Omer Levy, Veselin Stoyanov, and Luke Zettlemoyer. 2020{\natexlab{a}}.
\newblock \href {https://doi.org/10.18653/v1/2020.acl-main.703} {Bart:
  Denoising sequence-to-sequence pre-training for natural language generation,
  translation, and comprehension}.
\newblock In \emph{Proceedings of the 58th Annual Meeting of the Association
  for Computational Linguistics}, pages 7871--7880.

\bibitem[{Lewis et~al.(2020{\natexlab{b}})Lewis, Perez, Piktus, Petroni,
  Karpukhin, Goyal, K\"{u}ttler, Lewis, Yih, Rockt\"{a}schel, Riedel, and
  Kiela}]{lewis2020retrieval}
Patrick Lewis, Ethan Perez, Aleksandra Piktus, Fabio Petroni, Vladimir
  Karpukhin, Naman Goyal, Heinrich K\"{u}ttler, Mike Lewis, Wen-tau Yih, Tim
  Rockt\"{a}schel, Sebastian Riedel, and Douwe Kiela. 2020{\natexlab{b}}.
\newblock \href
  {https://proceedings.neurips.cc/paper/2020/file/6b493230205f780e1bc26945df7481e5-Paper.pdf}
  {Retrieval-augmented generation for knowledge-intensive nlp tasks}.
\newblock In \emph{Advances in Neural Information Processing Systems}, pages
  9459--9474.

\bibitem[{Lian et~al.(2019)Lian, Xie, Wang, Peng, and Wu}]{lian2019learning}
Rongzhong Lian, Min Xie, Fan Wang, Jinhua Peng, and Hua Wu. 2019.
\newblock \href {"https://www.ijcai.org/proceedings/2019/0706"} {Learning to
  select knowledge for response generation in dialog systems}.
\newblock In \emph{IJCAI International Joint Conference on Artificial
  Intelligence}, page 5081.

\bibitem[{Ma et~al.(2020)Ma, Zhang, Sun, and Liu}]{ma2020compare}
Longxuan Ma, Weinan Zhang, Runxin Sun, and Ting Liu. 2020.
\newblock \href {"https://aclanthology.org/2020.findings-emnlp.122"} {A compare
  aggregate transformer for understanding document-grounded dialogue}.
\newblock In \emph{Findings of the Association for Computational Linguistics:
  EMNLP 2020}, pages 1358--1367.

\bibitem[{Mao et~al.(2021)Mao, He, Liu, Shen, Gao, Han, and
  Chen}]{mao2021generation}
Yuning Mao, Pengcheng He, Xiaodong Liu, Yelong Shen, Jianfeng Gao, Jiawei Han,
  and Weizhu Chen. 2021.
\newblock \href {"https://aclanthology.org/2021.acl-long.316"}
  {Generation-augmented retrieval for open-domain question answering}.
\newblock In \emph{Proceedings of the 59th Annual Meeting of the Association
  for Computational Linguistics and the 11th International Joint Conference on
  Natural Language Processing (Volume 1: Long Papers)}, pages 4089--4100.

\bibitem[{Marcus(2020)}]{marcus2020next}
Gary Marcus. 2020.
\newblock \href {http://arxiv.org/abs/2002.06177} {The next decade in {AI}:
  four steps towards robust artificial intelligence}.
\newblock \emph{arXiv preprint arXiv:2002.06177}.

\bibitem[{Moghe et~al.(2018)Moghe, Arora, Banerjee, and
  Khapra}]{moghe2018towards}
Nikita Moghe, Siddhartha Arora, Suman Banerjee, and Mitesh~M Khapra. 2018.
\newblock \href {"https://aclanthology.org/D18-1255"} {Towards exploiting
  background knowledge for building conversation systems}.
\newblock In \emph{Proceedings of the 2018 Conference on Empirical Methods in
  Natural Language Processing}, pages 2322--2332.

\bibitem[{Ouyang et~al.(2022)Ouyang, Wu, Jiang, Almeida, Wainwright, Mishkin,
  Zhang, Agarwal, Slama, Ray et~al.}]{ouyang2022training}
Long Ouyang, Jeff Wu, Xu~Jiang, Diogo Almeida, Carroll~L Wainwright, Pamela
  Mishkin, Chong Zhang, Sandhini Agarwal, Katarina Slama, Alex Ray, et~al.
  2022.
\newblock \href {https://arxiv.org/abs/2203.02155} {Training language models to
  follow instructions with human feedback}.
\newblock \emph{arXiv preprint arXiv:2203.02155}.

\bibitem[{Paranjape et~al.(2022)Paranjape, Khattab, Potts, Zaharia, and
  Manning}]{paranjape2022hindsight}
Ashwin Paranjape, Omar Khattab, Christopher Potts, Matei Zaharia, and
  Christopher~D Manning. 2022.
\newblock \href {https://openreview.net/forum?id=Vr_BTpw3wz} {Hindsight:
  Posterior-guided training of retrievers for improved open-ended generation}.
\newblock In \emph{International Conference on Learning Representations}.

\bibitem[{Petroni et~al.(2021)Petroni, Piktus, Fan, Lewis, Yazdani, De~Cao,
  Thorne, Jernite, Karpukhin, Maillard et~al.}]{petroni2021kilt}
Fabio Petroni, Aleksandra Piktus, Angela Fan, Patrick Lewis, Majid Yazdani,
  Nicola De~Cao, James Thorne, Yacine Jernite, Vladimir Karpukhin, Jean
  Maillard, et~al. 2021.
\newblock \href {https://doi.org/10.18653/v1/2021.naacl-main.200} {Kilt: a
  benchmark for knowledge intensive language tasks}.
\newblock In \emph{Proceedings of the 2021 Conference of the North American
  Chapter of the Association for Computational Linguistics: Human Language
  Technologies}, pages 2523--2544.

\bibitem[{Petroni et~al.(2019)Petroni, Rockt{\"a}schel, Riedel, Lewis, Bakhtin,
  Wu, and Miller}]{petroni2019language}
Fabio Petroni, Tim Rockt{\"a}schel, Sebastian Riedel, Patrick Lewis, Anton
  Bakhtin, Yuxiang Wu, and Alexander Miller. 2019.
\newblock \href {https://aclanthology.org/D19-1250} {Language models as
  knowledge bases?}
\newblock In \emph{Proceedings of the 2019 Conference on Empirical Methods in
  Natural Language Processing and the 9th International Joint Conference on
  Natural Language Processing (EMNLP-IJCNLP)}, pages 2463--2473.

\bibitem[{Raffel et~al.(2020)Raffel, Shazeer, Roberts, Lee, Narang, Matena,
  Zhou, Li, and Liu}]{raffel2020exploring}
Colin Raffel, Noam Shazeer, Adam Roberts, Katherine Lee, Sharan Narang, Michael
  Matena, Yanqi Zhou, Wei Li, and Peter~J Liu. 2020.
\newblock \href {http://jmlr.org/papers/v21/20-074.html} {Exploring the limits
  of transfer learning with a unified text-to-text transformer}.
\newblock \emph{Journal of Machine Learning Research}, 21(140):1--67.

\bibitem[{Raposo et~al.(2022)Raposo, Ribeiro, Martins, and
  Coheur}]{raposo2022question}
Gon{\c{c}}alo Raposo, Rui Ribeiro, Bruno Martins, and Lu{\'\i}sa Coheur. 2022.
\newblock \href {https://doi.org/10.1007/978-3-030-99739-7_23} {Question
  rewriting? assessing its importance for conversational question answering}.
\newblock In \emph{European Conference on Information Retrieval}, pages
  199--206.

\bibitem[{Ren et~al.(2021)Ren, Qu, Liu, Zhao, She, Wu, Wang, and
  Wen}]{ren2021rocketqav2}
Ruiyang Ren, Yingqi Qu, Jing Liu, Wayne~Xin Zhao, Qiaoqiao She, Hua Wu, Haifeng
  Wang, and Ji-Rong Wen. 2021.
\newblock \href {https://aclanthology.org/2021.emnlp-main.224} {Rocketqav2: A
  joint training method for dense passage retrieval and passage re-ranking}.
\newblock In \emph{Proceedings of the 2021 Conference on Empirical Methods in
  Natural Language Processing}, pages 2825--2835.

\bibitem[{Roller et~al.(2021)Roller, Dinan, Goyal, Ju, Williamson, Liu, Xu,
  Ott, Shuster, Smith, Boureau, and Weston}]{roller2020recipes}
Stephen Roller, Emily Dinan, Naman Goyal, Da~Ju, Mary Williamson, Yinhan Liu,
  Jing Xu, Myle Ott, Kurt Shuster, Eric~M Smith, Y-Lan Boureau, and Jason
  Weston. 2021.
\newblock \href {https://www.aclweb.org/anthology/2021.eacl-main.24} {Recipes
  for building an open-domain chatbot}.
\newblock In \emph{Proceedings of the 16th Conference of the European Chapter
  of the Association for Computational Linguistics}.

\bibitem[{Shuster et~al.(2021)Shuster, Poff, Chen, Kiela, and
  Weston}]{shuster2021retrieval}
Kurt Shuster, Spencer Poff, Moya Chen, Douwe Kiela, and Jason Weston. 2021.
\newblock \href {"https://aclanthology.org/2021.findings-emnlp.320"} {Retrieval
  augmentation reduces hallucination in conversation}.
\newblock In \emph{Findings of the Association for Computational Linguistics:
  EMNLP 2021}, pages 3784--3803.

\bibitem[{Thulke et~al.(2021)Thulke, Daheim, Dugast, and
  Ney}]{thulke2021efficient}
David Thulke, Nico Daheim, Christian Dugast, and Hermann Ney. 2021.
\newblock \href {"https://arxiv.org/abs/2102.04643"} {Efficient retrieval
  augmented generation from unstructured knowledge for task-oriented dialog}.
\newblock \emph{arXiv preprint arXiv:2102.04643}.

\bibitem[{Tian et~al.(2022)Tian, Lin, Song, Bao, Wang, He, Sun, and
  Wu}]{tian2022q}
Xin Tian, Yingzhan Lin, Mengfei Song, Siqi Bao, Fan Wang, Huang He, Shuqi Sun,
  and Hua Wu. 2022.
\newblock \href {http://arxiv.org/abs/2210.07564} {Q-tod: A query-driven
  task-oriented dialogue system}.
\newblock \emph{arXiv preprint arXiv:2210.07564}.

\bibitem[{Vakulenko et~al.(2021)Vakulenko, Longpre, Tu, and
  Anantha}]{vakulenko2021question}
Svitlana Vakulenko, Shayne Longpre, Zhucheng Tu, and Raviteja Anantha. 2021.
\newblock \href {https://doi.org/10.1145/3437963.3441748} {Question rewriting
  for conversational question answering}.
\newblock In \emph{Proceedings of the 14th ACM International Conference on Web
  Search and Data Mining}, pages 355--363.

\bibitem[{Wen et~al.(2017)Wen, Vandyke, Mrk{\v{s}}i{\'c}, Gasic, Barahona, Su,
  Ultes, and Young}]{wen2017network}
Tsung-Hsien Wen, David Vandyke, Nikola Mrk{\v{s}}i{\'c}, Milica Gasic, Lina
  M~Rojas Barahona, Pei-Hao Su, Stefan Ultes, and Steve Young. 2017.
\newblock \href {"https://aclanthology.org/E17-1042"} {A network-based
  end-to-end trainable task-oriented dialogue system}.
\newblock In \emph{Proceedings of the 15th Conference of the European Chapter
  of the Association for Computational Linguistics: Volume 1, Long Papers},
  pages 438--449.

\bibitem[{Wu et~al.(2021)Wu, Luan, Rashkin, Reitter, and Tomar}]{wu2021conqrr}
Zeqiu Wu, Yi~Luan, Hannah Rashkin, David Reitter, and Gaurav~Singh Tomar. 2021.
\newblock \href {https://arxiv.org/abs/2112.08558} {Conqrr: Conversational
  query rewriting for retrieval with reinforcement learning}.
\newblock \emph{arXiv preprint arXiv:2112.08558}.

\bibitem[{Xie et~al.(2022)Xie, Wu, Shi, Zhong, Scholak, Yasunaga, Wu, Zhong,
  Yin, Wang et~al.}]{xie2022unifiedskg}
Tianbao Xie, Chen~Henry Wu, Peng Shi, Ruiqi Zhong, Torsten Scholak, Michihiro
  Yasunaga, Chien-Sheng Wu, Ming Zhong, Pengcheng Yin, Sida~I Wang, et~al.
  2022.
\newblock \href {http://arxiv.org/abs/2201.05966} {Unifiedskg: Unifying and
  multi-tasking structured knowledge grounding with text-to-text language
  models}.
\newblock \emph{arXiv preprint arXiv:2201.05966}.

\bibitem[{Yu et~al.(2020)Yu, Liu, Yang, Xiong, Bennett, Gao, and
  Liu}]{yu2020few}
Shi Yu, Jiahua Liu, Jingqin Yang, Chenyan Xiong, Paul Bennett, Jianfeng Gao,
  and Zhiyuan Liu. 2020.
\newblock \href {https://doi.org/10.1145/3397271.3401323} {Few-shot generative
  conversational query rewriting}.
\newblock In \emph{Proceedings of the 43rd International ACM SIGIR conference
  on research and development in Information Retrieval}, pages 1933--1936.

\bibitem[{Zhao et~al.(2020{\natexlab{a}})Zhao, Wu, Xu, Tao, Zhao, and
  Yan}]{zhao2020knowledge}
Xueliang Zhao, Wei Wu, Can Xu, Chongyang Tao, Dongyan Zhao, and Rui Yan.
  2020{\natexlab{a}}.
\newblock \href {"https://aclanthology.org/2020.emnlp-main.272"}
  {Knowledge-grounded dialogue generation with pre-trained language models}.
\newblock In \emph{Proceedings of the 2020 Conference on Empirical Methods in
  Natural Language Processing (EMNLP)}, pages 3377--3390.

\bibitem[{Zhao et~al.(2020{\natexlab{b}})Zhao, Wu, and Xu}]{zhao2020pre}
Yufan Zhao, Wei Wu, and Can Xu. 2020{\natexlab{b}}.
\newblock \href {"https://arxiv.org/abs/2011.09708"} {Are pre-trained language
  models knowledgeable to ground open domain dialogues?}
\newblock \emph{arXiv preprint arXiv:2011.09708}.

\bibitem[{Zhou et~al.(2018)Zhou, Prabhumoye, and Black}]{zhou2018dataset}
Kangyan Zhou, Shrimai Prabhumoye, and Alan~W Black. 2018.
\newblock \href {"https://aclanthology.org/D18-1076"} {A dataset for document
  grounded conversations}.
\newblock In \emph{Proceedings of the 2018 Conference on Empirical Methods in
  Natural Language Processing}, pages 708--713.

\end{thebibliography}
\bibliographystyle{acl_natbib}

% \clearpage
% \newpage
\appendix

\section{Model Details}
\label{sec:appendixHyper}
We apply iterative training of our model with an inner-outer loop structure several times until convergence. We used 8 NVIDIA A100 GPUs with approximately 4 hours for each iteration.

The outer loop executes query generation and knowledge selection to collect training data. 
Given a query for QReCC and WoW, we retrieve top-50 knowledge from the knowledge base and get the top-1 after reranking. For SMD, we obtain top-3 knowledge after reranking due to the requirement of multiple knowledge for response generation. 

The inner loop updates the model with collected data. The hyperparameters are the same for all datasets but differentiate the learning rate by model scale, detailed in Table \ref{tab:hyper}. The model checkpoint is determined by the F1 score in the validation set.

\begin{table}[h]
\begin{center}
\small
\renewcommand{\arraystretch}{1.1}
\begin{tabular}{@{}lp{0.12\textwidth}<{\centering}p{0.12\textwidth}<{\centering}@{}}
\toprule
\multirow{2}{*}{Parameters}  & \multicolumn{2}{c}{Model Scale} \\
&    Base   &   Large          \\ \midrule
Optimizer &   AdamW   &   AdamW       \\
Learning Rate  &   5e-5   &    1e-5  \\
LR Scheduler   &   Linear   &    Linear  \\ 
Batch Size  &   16   &    16  \\
Inner Epoch  &   2   &    2   \\
Input Length  &   1024   &    1024   \\
Output Length  &   128   &    128 
\\\bottomrule
\end{tabular}
\end{center}
\caption{Hyperparameters used in QReCC, SMD, and WoW.}
\label{tab:hyper}
\end{table}

\section{Scoring Criteria in Human Evaluation}
The criteria of human evaluation are provided in Table \ref{tab:criteria}.

\begin{table}[ht]
\begin{center}
\small
\renewcommand{\arraystretch}{1.1}
\begin{tabular}{@{}p{0.04\textwidth}<{\centering} p{0.41\textwidth}@{}}
\toprule
Score & \multicolumn{1}{c}{Coherence}                \\ \midrule
0 &
  \begin{tabular}[c]{@{}l@{}}- The response is not related with the context.\\ - The response simply repeats the context.\\ - The response has obvious conflicts with context.\\ - There are serious logic conflicts within response.\end{tabular} \\
\midrule
1 &
  \begin{tabular}[c]{@{}l@{}}- The response has minor conflicts with the context.\\ - There are some minor logic conflicts in response.\end{tabular} \\
\midrule
2     & - The response is coherent with the context. \\ \bottomrule
\\

\toprule
Score & \multicolumn{1}{c}{Groundedness}       \\ 
\midrule
0 &
\begin{tabular}[c]{@{}l@{}}- The response contains no information. \\ - The response simply repeats the context and con-\\tains no additional information.\end{tabular} \\
\midrule
1     & - The response contains a little additional information. \\ 
\midrule
2     & - The response has appropriate information. \\ 
\bottomrule
\\

\toprule
Score & \multicolumn{1}{c}{Engagingness}\\ 
\midrule
0     & - I don’t want to talk with this system.   \\
\midrule
1     & - It is kind of boring, but it is still ok to talk with this system. \\
\midrule
2     & - I would like to talk with this system for a long conversation.     \\ \bottomrule
\end{tabular}
\end{center}
\caption{Score details of metrics in human evaluation.}
\label{tab:criteria}
\end{table}

\section{Model Scalability}
\label{sec:scale}

\begin{figure}[h]
	\centering
	\includegraphics[width=0.48\textwidth]{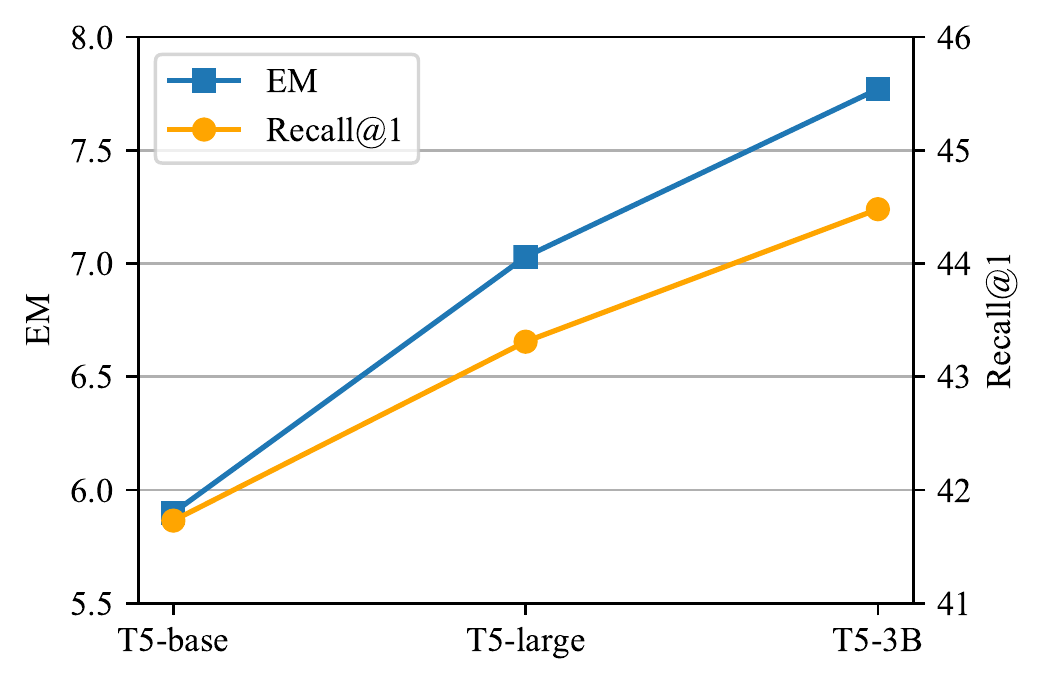}
	\caption{Effects of model scaling on QReCC test set.}
	\label{fig:scale}
\end{figure}

Motivated by the generally observed phenomenon that the generation ability improves with the model size, we evaluate the scalability of {\M} on the QReCC dataset with T5-base, T5-large, and T5-3B. The metrics of EM and Recall@1 are criteria to evaluate response generation and query generation, respectively. As shown in Figure \ref{fig:scale}, the EM scores of generated response increase by roughly 0.9\% with each scale-up, and Recall@1 scores of generated query experience a 1.4\% average boost for each scale-up. Specifically, there is a more significant benefit when increasing the model size from T5-base to T5-large than T5-large to T5-3B. Furthermore, as the improved knowledge selection also contributes to response generation, the EM scores have a more notable relative increase (+16.4\%) compared to the Recall@1 score (+3.4\%). 

\section{Few Query Supervision}
\label{sec:fewshot}
\begin{figure}[h]
	\centering
	\includegraphics[width=0.48\textwidth]{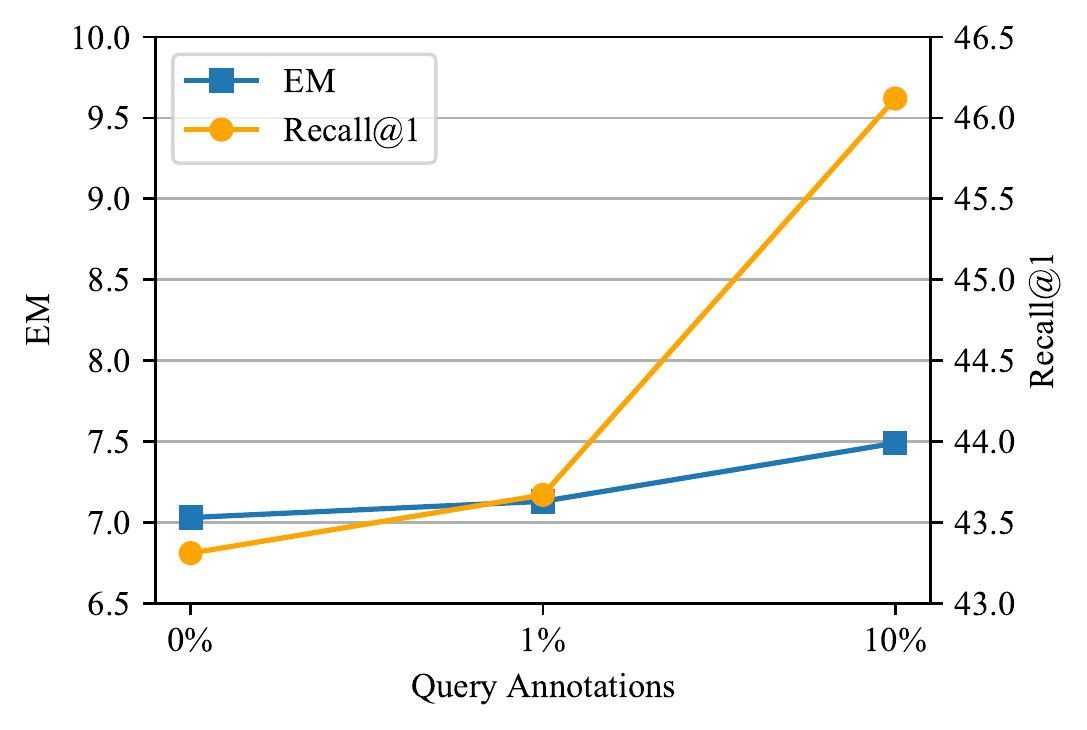}
	\caption{Performance of {\M} in the zero-shot and few-shot settings of QReCC.}
	\label{fig:fewshot}
\end{figure}

{\M} has limitations in resolving complex query conditions. To bridge the gaps, we incorporate a few query annotations into training. To be specific, 1\% or 10\% of human-rewritten queries replace the context-sensitive guidance during training to regulate query generation and facilitate joint training. Figure \ref{fig:fewshot} shows that some query annotations can further improve query generation and response generation, especially with more supervised data. 
It is worth noting that the marginal benefit of knowledge selection on response generation is relatively small in models of the same scale.
According to the examples in Table \ref{tab:fewshotCase}, adding 1\% supervised data has a minor impact on the queries, while adding 10\% supervised data enables the model to rewrite the last utterance without impairing its original ability to extract previous contexts.

\begin{table*}[t]
\setlength\tabcolsep{4pt}
\begin{center}
\small
\renewcommand{\arraystretch}{1.1}
\begin{tabular}{@{}l p{0.8\textwidth}@{}}
\toprule
\multicolumn{2}{@{}l}{Cherry picked example} \\ \midrule
Dialogue History & \makecell[tl]{\textbf{User:} what were some of \textcolor{my_blue}{\textit{john stossel}}'s most popular publications? \\ 
\textbf{System:} \textcolor{my_blue}{\textit{give me a break: how i exposed hucksters, cheats, and scam artists and became the scourge}} \\
\textcolor{my_blue}{\textit{of the liberal media}} is an autobiography documenting stossel's career and philosophical transition.\\
...  \\
\textbf{User:} what was the response?} \\
{\M} Query  & give me a break: how i exposed hucksters, cheats, and scam artists and became the scourge of the liberal media \textbf{received generally positive reviews from critics}. (\usym{1F5F8})\\
{\M} (1\%) Query  & give me a break: how i exposed hucksters, cheats, and scam artists and became the scourge of the liberal media \textbf{received mixed reviews from critics}. (\usym{1F5F8}) \\
{\M} (10\%) Query & what was the response to give me a break: how i exposed hucksters, cheats, and scam artists and became the scourge of the liberal media \textbf{by john stossel}? (\usym{1F5F8}) \\
Golden Query & what was the response to give me a break: how i exposed hucksters, cheats, and scam artists and became the scourge of the liberal media?
\\ \midrule
\multicolumn{2}{@{}l}{Lemon picked example} \\ \midrule
Dialogue History & \makecell[tl]{\textbf{User:} what part did \textcolor{my_blue}{\textit{victor mclaglen}} play in \textcolor{my_blue}{\textit{happy days}}?  \\ 
\textbf{System:} victor mclaglen was a minstrel show performer in the film, happy days. \\
...  \\
\textbf{User:} what other films did he play in?} \\
{\M} Query  & victor mclaglen was a minstrel show performer in the film, happy days. (\usym{2717}) \\
{\M} (1\%) Query  & victor mclaglen was a minstrel show performer in the film, happy days. (\usym{2717})  \\
{\M} (10\%) Query & what other films did \textbf{victor mclaglen} play in \textbf{besides happy days}? victor mclaglen was a minstrel show performer in the film, happy days. (\usym{1F5F8})  \\
Golden Query & what other speaking films did victor mclaglen play in besides happy days?
\\
\bottomrule
\end{tabular}
\end{center}
\caption{Examples of queries generated by {\M} on QReCC test set. All examples are uncased. 1\% and 10\% following {\M} mean the proportion of query annotations used in the few-shot setting. \textcolor{my_blue}{\textit{Blue}} indicates the helpful information in the dialogue context, and \textbf{Bold} highlights the main difference between queries. \usym{1F5F8}/\usym{2717} inside the bracket indicates top-1 selection accuracy.}
\label{tab:fewshotCase}
\end{table*}

\section{Additional Qualitative Results}
\label{sec:appendixCase}
The following tables provide qualitative results of models in Table \ref{tab:main} for all datasets. For query generation, Table \ref{tab:smd} and Table \ref{tab:wow} contain examples of SMD and WoW where the generated queries also support the heterogeneous query generation modality. The query generator of {\M} tends to extract relevant information from the dialogue context or generate a plausible response-looking query interfusing the essential information from the dialogue context. The response-looking query guides knowledge selection toward the target response, while the authentic information derived from the dialogue context ensures the relevance of knowledge selection. For response generation, the generated response on all datasets exemplifies our model's superior knowledge utilization ability. The selected knowledge passages are lengthy on QReCC and WoW, while top-3 knowledge is selected on SMD. Therefore, the knowledge inevitably includes useless information for reply. However, {\M} is capable of denoising and locating the appropriate span within the long knowledge passage.

\begin{table}[h]
\setlength\tabcolsep{2pt}
\begin{center}
\small
\renewcommand{\arraystretch}{1.1}
\begin{tabular}{@{}lp{0.345\textwidth}@{}}
\toprule
\makecell[tl]{Dialogue \\ history} & \makecell[tl]{\textbf{User}: when did the tv show the rifleman air \\
\textbf{System}: \textcolor{my_blue}{\textit{the rifleman aired from september}} \\ \textcolor{my_blue}{\textit{30, 1958, to april 8, 1963.}} \\
\textbf{User}: in what network} \\
{\M} Query &   the rifleman aired from september 30, 1958, to april 8, 1963 on nbc    \\
\makecell[tl]{Selected \\Knowledge} &  the rifleman - wikipedia ... the rifleman \underline{aired on abc} from september 30, 1958, to april 8, 1963, as a production of four star television ...  \\
{\M} Reply &   the rifleman aired on abc\\ \midrule
\makecell[tl]{Dialogue \\ history} & \makecell[tl]{\textbf{User}: when did the \textcolor{my_blue}{\textit{milwaukee bucks}} win \\the nba championship \\
\textbf{System}: the milwaukee bucks won an nba \\ championship in 1971. \\
\textbf{User}: who was the team's \textcolor{my_blue}{\textit{coach in 1971}}} \\
{\M} Query &   the milwaukee bucks's coach in 1971 was john mccartney.    \\
\makecell[tl]{Selected \\Knowledge} & \makecell[tl]{List of Milwaukee Bucks head coaches ... \\
\underline{The franchise won its only NBA champion-} \\
\underline{ship ... Larry Costello} ...}  \\
{\M} Reply &   the milwaukee bucks won its only nba championship in the 1971 nba finals under the leadership of its first coach, larry costello.
\\ \bottomrule
\end{tabular}
\end{center}
\caption{Examples of QReCC test set. \textcolor{my_blue}{\textit{Blue}} marks the provenance of queries, and the underline indicates the source of response.}
\label{tab:qrecc}
\end{table}

\begin{table}[h]
\setlength\tabcolsep{2pt}
\begin{center}
\small
\renewcommand{\arraystretch}{1.1}
\begin{tabular}{@{}lp{0.345\textwidth}@{}}
\toprule
\makecell[tl]{Dialogue \\ history} & \makecell[tl]{\textbf{User}: what is the forecast for the next 7 da-\\ys \\
\textbf{System}: for what city are you interested in? \\
\textbf{User}: \textcolor{my_blue}{\textit{cleveland, will it be windy?}}} \\
{\M} Query &   the forecast for cleveland, will it be windy?    \\
\makecell[tl]{Selected \\Knowledge} & \makecell[tl]{\underline{cleveland, stormy, dry, stormy, hail, warm,} \\ \underline{hot, stormy} \\
boston, foggy, snow, windy, blizzard, clear \\ skies, windy, overcast \\
oakland, windy, hail, overcast, dry, hot, hail\\, clear skies}  \\
{\M} Reply &   it will not be windy in cleveland this week.\\ \midrule
\makecell[tl]{Dialogue \\ history} & \makecell[tl]{\textbf{User}: what time is my doctor apartment \\ and  who is it with? \\
\textbf{System}: are you talking about the  \textcolor{my_blue}{\textit{doctor}}\\\textcolor{my_blue}{\textit{appointment on wednesday}} or the one on \\the 5th? \\
\textbf{User}: the one on wednesday.}  \\
{\M} Query &   your doctor apartment on wednesday is at 1 pm.    \\
\makecell[tl]{Selected \\Knowledge} & \makecell[tl]{\underline{7pm, wednesday, father, doctor appointment} \\
6pm, the 5th, Alex, doctor appointment \\
2pm, thursday, Jon, dentist appointment} \\
{\M} Reply &   your doctor appointment on wednesday is at 7pm with your father.
\\ \bottomrule
\end{tabular}
\end{center}
\caption{Examples of SMD test set. \textcolor{my_blue}{\textit{Blue}} marks the provenance of queries, and the underline indicates the source of response.}
\label{tab:smd}
\end{table}

\begin{table}[h]
\setlength\tabcolsep{2pt}
\begin{center}
\small
\renewcommand{\arraystretch}{1.1}
\begin{tabular}{@{}lp{0.345\textwidth}@{}}
\toprule
\makecell[tl]{Dialogue \\ history} & \makecell[tl]{\textbf{User}: I own a Ferrari, do you know  anyth-\\ing about them? \\
\textbf{System}: Yes! \textcolor{my_blue}{\textit{Ferrari is a company based}} \\ \textcolor{my_blue}{\textit{in Maranello, Italy}} \\
\textbf{User}: \textcolor{my_blue}{\textit{I see, who founded Ferrari?}}} \\
{\M} Query &   ferrari is a company based in maranello, italy i see, who founded ferrari?    \\
\makecell[tl]{Selected \\Knowledge} & Ferrari (; ) is an Italian luxury sports car manufacturer based in Maranello. \underline{Founded by Enzo Ferrari} in 1939  ...   \\ 
{\M} Reply &   Ferrari was founded by Enzo Ferrari in 1939 \\ \midrule
Dialogue history & \makecell[tl]{\textbf{User}: My mother always enjoyed \textcolor{my_blue}{\textit{jazz mu-}} \\ \textcolor{my_blue}{\textit{sic}}. I might try to find a jazz concert to \\ give it a try} \\
{\M} Query &   jazz music is a genre of music that originated in New Orleans.    \\
\makecell[tl]{Selected \\Knowledge} & \makecell[tl]{\underline{Jazz is a music genre that originated in the} \\
\underline{African-American communities of New O-} \\
\underline{rleans, United States} ...}  \\
{\M} Reply &   Jazz is a music genre that originated in the African-American communities of New Orleans
\\ \bottomrule
\end{tabular}
\end{center}
\caption{Examples of WoW dev set. \textcolor{my_blue}{\textit{Blue}} marks the provenance of queries, and the underline indicates the source of response.}
\label{tab:wow}
\end{table}

\end{document}